\documentclass[10pt,twocolumn,letterpaper]{article}

\usepackage{cvpr}
\usepackage{times}
\usepackage{epsfig}
\usepackage{graphicx}
\usepackage{amsmath}
\usepackage{amssymb}
\usepackage{booktabs}
\usepackage[numbers,sort]{natbib} 

\usepackage[pagebackref=true,breaklinks=true,letterpaper=true,colorlinks,bookmarks=false]{hyperref}

\cvprfinalcopy 

\def\cvprPaperID{****} 

\newcommand{\datasetname}{{\tt RareAct}}
\newcommand{\numclip}{7607 }
\newcommand{\numvideo}{905 }
\newcommand{\projecturl}{\url{https://github.com/antoine77340/RareAct}}
\ifcvprfinal\fi
\begin{document}

\title{\datasetname: A video dataset of unusual interactions}


\author{
	Antoine Miech\textsuperscript{1}
	\quad\quad\quad
	Jean-Baptiste Alayrac
	\\
		\quad\quad
	Ivan Laptev\textsuperscript{1}
	\quad\quad\quad\quad\quad
	Josef Sivic\textsuperscript{1,2}
	\quad\quad\quad
	Andrew Zisserman\textsuperscript{3}
	\\
	\small{$^1$ENS/Inria \small \quad $^2$CIIRC CTU \quad $^3$VGG Oxford}
	\\
	\small{\texttt{antoine77340@hotmail.com}}
	\\
	\small{\projecturl}
}

\maketitle
\pagestyle{plain}

\begin{abstract}
This paper introduces a manually annotated video dataset of unusual actions, namely \datasetname, including actions such as \textit{`blend phone'}, \textit{`cut keyboard'} and \textit{`microwave shoes'}.
\datasetname{} aims at evaluating the zero-shot and few-shot compositionality of action recognition models for unlikely compositions of common action verbs and object nouns.
It contains 122 different actions which were obtained by combining verbs and nouns rarely co-occurring together in the large-scale textual corpus from HowTo100M~\cite{miech19howto100m}, but that frequently appear separately.
We provide benchmarks using a state-of-the-art HowTo100M pretrained video and text model and show that zero-shot and few-shot compositionality of actions remains a challenging and unsolved task.
The dataset is publicly available for download at \projecturl.
\end{abstract}

%

\section{Introduction}

\begin{figure}[t]
\centering
	\includegraphics[width=\columnwidth]{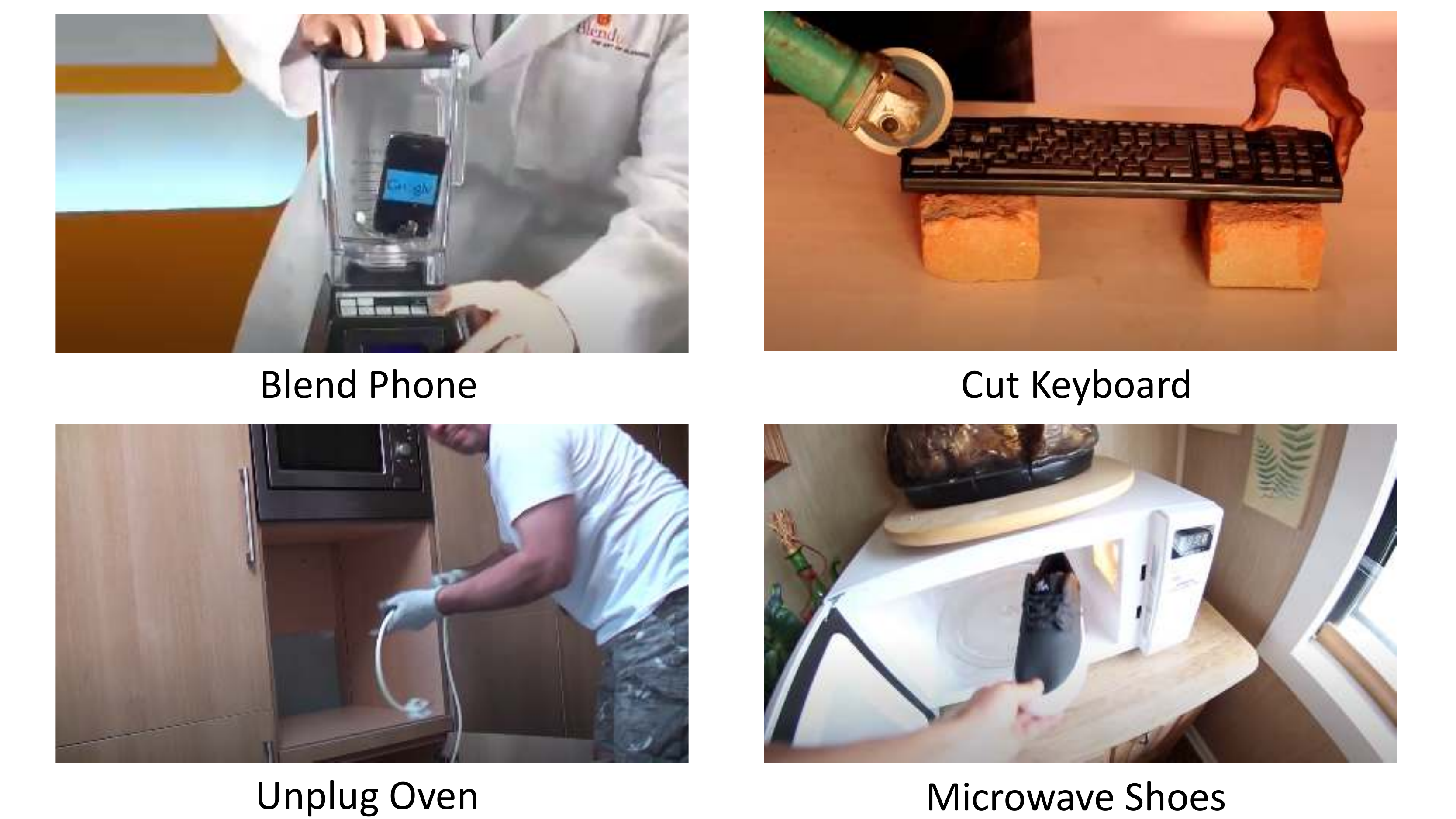}
\caption{\label{fig:teaser} Examples of collected action verbs and object nouns compositions from \datasetname.
}
\end{figure}

Many human actions involve interacting with objects.
These actions can often be decomposed into an action verb followed by an object noun (e.g.\ cut paper, cut tree, fold paper).
Many of the popular action recognition datasets such as Kinetics \cite{kay2017kinetics} or AVA \cite{gu2018ava} concentrate on frequently occurring actions, to obtain a large number of clips for each action class.
In contrast, we aim at providing a benchmark for evaluating the compositionality of action recognition models for {\em rare}  human-object interactions.
To this end, we introduce a manually annotated video dataset of rare actions, \datasetname, with unlikely compositions of common action verbs and object nouns such as: \textit{blend phone}, \textit{cut keyboard}, \textit{unplug oven} or \textit{microwave shoes} as illustrated in Figure \ref{fig:teaser} (the full list is provided in section \ref{dataset}).
To correctly assess compositionality, we make sure to collect hard negatives examples that either share the same action verb or object noun for each action class.
The taxonomy of \datasetname{} is constructed by collecting rarely co-occurring action verbs and object nouns from the large textual corpus of HowTo100M~\cite{miech19howto100m}.
We emphasize that this dataset is only an \emph{evaluation} dataset notably meant to be used to evaluate models trained on the HowTo100M dataset.

\paragraph{Related work.}
Our work is inspired by UnRel \cite{Peyre17}, which is an image dataset composed of unusual spatial relations triplets such as (elephant, ride, bike). 
Each element within the triplets are common noun or verbs, easily recognizable alone but challenging to identify when combined as these triplets are rarely or never seen at training.
Our work instead, focuses on the compositionality aspect of human action involving objects rather than spatial relations of objects in images. Our \datasetname{} dataset follows a line of datasets of images and videos of unusual situations such as:
out-of-context objects~\cite{Choi2012ContextMA};
dangerous, but rare pedestrian scenes in the `Precarious Pedestrians' dataset~\cite{Huang_2017_CVPR};
and unintentional actions in videos in the 	`OOPS!'  dataset~\cite{Epstein_2020_CVPR}.

The EPIC-KITCHENS video dataset \cite{damen2018scaling} is the closest {\em video} dataset related to ours, where actions are also annotated as a combination of a verb and a noun.
A few combinations of verbs and nouns are also rarely or never seen at training. Their work, however, focuses on ego-centric videos in the cooking domain.
In contrast, \datasetname{} is not constrained to either ego-centric videos or cooking videos and fully focuses on the rare composition of human-object actions.
\newline

\begin{figure}[t]
\centering
	\includegraphics[width=\columnwidth]{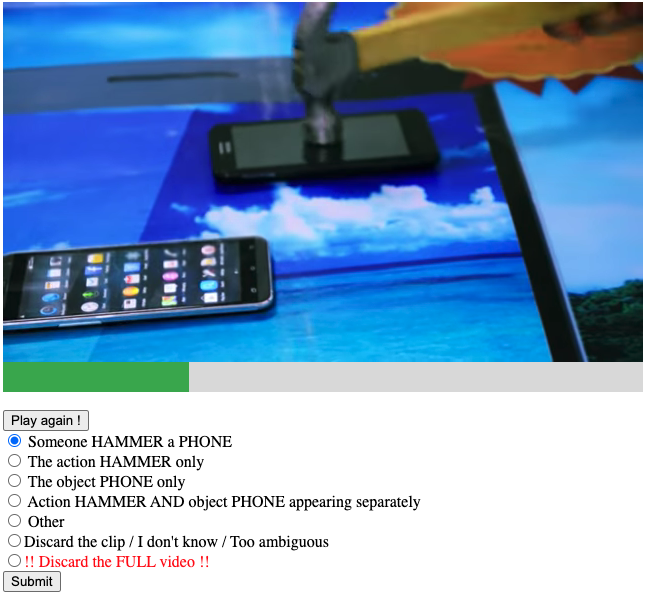}
\caption{\label{fig:annotation_tool} Screenshot from the annotation tool used to annotate the video clips. The annotator is asked to select between 7 choices while the 10-seconds video is continuously looping.
}
\end{figure}

\paragraph{Paper outline.}
We explain in section \ref{dataset} the collection and annotation process and provide some statistics about \datasetname.
In section \ref{metric}, we provide useful metrics that are suited to assess the compositionality of action recognition models on \datasetname.
Next, in section \ref{benchmark}, we provide several benchmarks based on a state-of-the-art text and video model \cite{miech2020end} trained on HowTo100M.

\section{Dataset}
\label{dataset}

\paragraph{Collection and annotation.}

To assess the compositionality aspect of action recognition models, we
consider combinations of verbs and nouns that satisfy the following two criteria:
\textit{(i)} The verbs and nouns never or rarely co-occur together in the HowTo100M~\cite{miech19howto100m} textual corpus; and
\textit{(ii)} The verbs and nouns frequently appear separately in the HowTo100M~\cite{miech19howto100m} to ensure that there is no challenge in either recognizing the action verb or the object noun.

Given this taxonomy, we search on a popular video-sharing platform, the top-ranked videos (disjoint from HowTo100M) using the verb followed by the noun as a query.
We then split each video into 10 seconds contiguous clips for annotation.
For each 10-second video clip, we ask annotators to choose between several options as illustrated in Figure~\ref{fig:annotation_tool}.
We give the meaning of each option next along with a precise example for the (verb, noun) pair (hammer, phone):
\begin{enumerate}
    \item Verb is applied to Noun (positive example): \small{\eg we see someone hammering a phone}. 
    \item Only the Verb is seen (hard negative example): \small{\eg we see someone hammering a nail but no phone is visible}. 
    \item Only the Noun is seen (hard negative example): \small{\eg we see a phone but the action of hammering is not happening}. 
    \item Verb and Noun are seen but Verb is not applied to the Noun (hard negative example): \small{\eg we see someone hammering a nail and there is also a phone visible in the clip}. 
    \item Neither apply (negative example): \small{\eg no phone nor the hammering action is happening}. 
    \item Too ambiguous / I don't know (example discarded): \small{\eg unclear case where the action is not clearly performed for example}. 
    \item Discard the full video (every example within the same video are discarded)
\end{enumerate}
Through this process, we categorize each video clip into positive, negative or hard negative examples for a given (verb, noun) action or eventually remove them from the dataset.
An important note is that it often happens that we obtain multiple positive clips (or negative) from the same original video.

\begin{figure*}[t]
\centering
	\includegraphics[width=\linewidth]{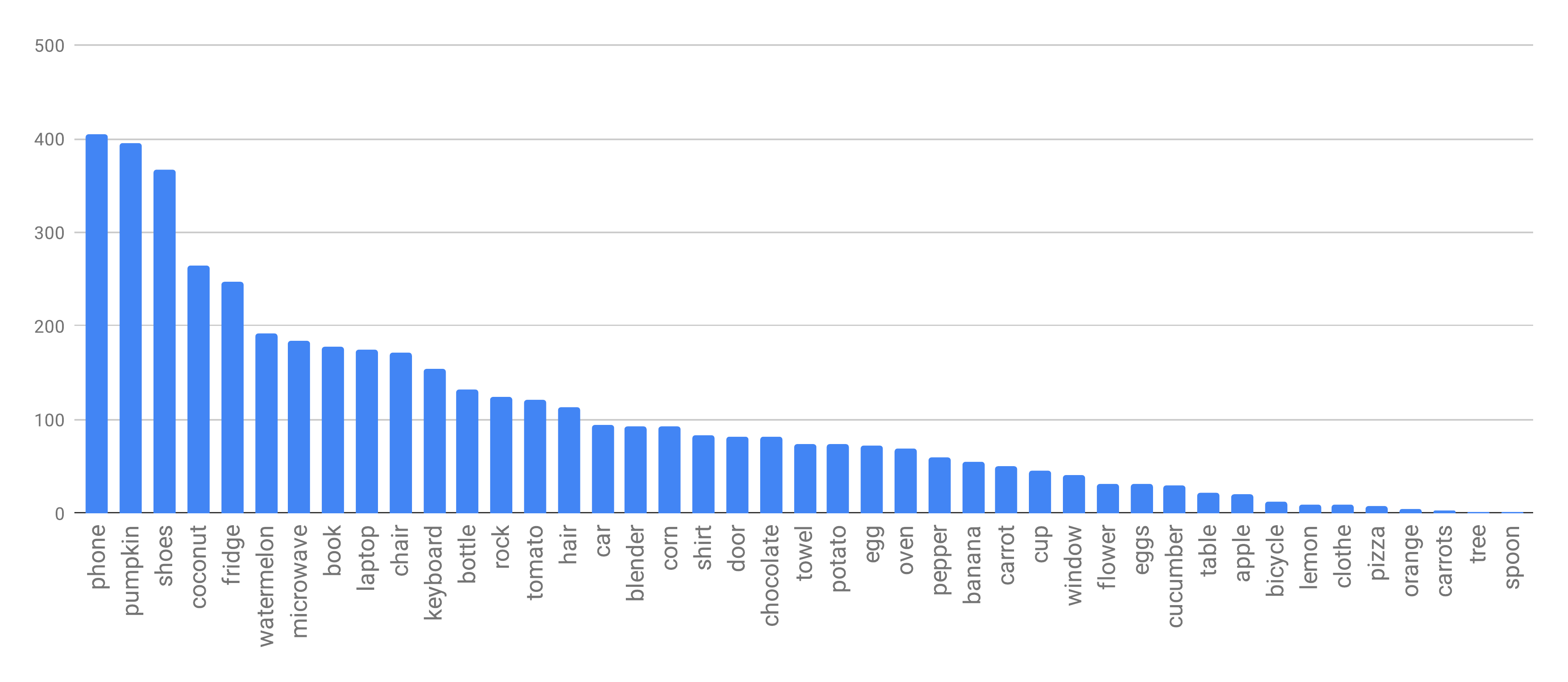}
\caption{\label{fig:nouns} Number of annotated example per noun.
}
\end{figure*}

\begin{figure*}[t]
\centering
	\includegraphics[width=0.92\linewidth]{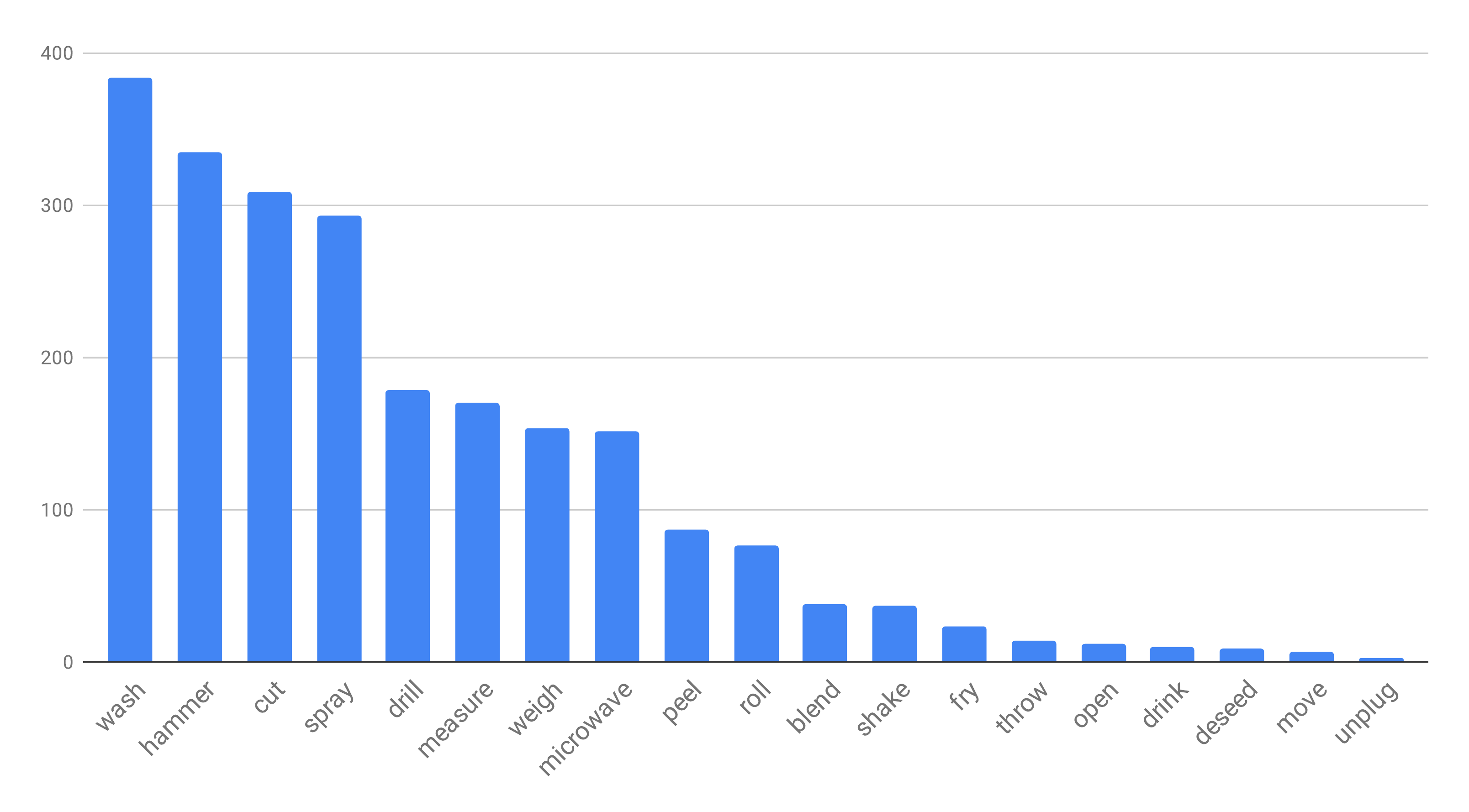}
\caption{\label{fig:verbs} Number of annotated example per verb.
}
\end{figure*}

\begin{table}[t]
\footnotesize
	\centering  
	\begin{tabular}{ccccccc}
	\toprule
	$\#$ video & $\#$ clip & $\#$ action & $\#$ positive   & $\#$ verb & $\#$ noun & Clip length   \\ \midrule
    \numvideo & \numclip & 122 & 1765 & 19 & 38            & 10 sec.          \\
\end{tabular}
\vspace{1mm}
	\caption{Main annotation statistics of \datasetname.}
	\label{table:stats}
\end{table}
\paragraph{Statistics.}
We provide the full taxonomy with the number of positive, negative and hard negative examples collected for each action class split in Table \ref{tab:tax1}, \ref{tab:tax2} and \ref{tab:tax3}.
Note we use the positive examples for a verb-noun pair as hard negatives for other actions sharing the same verb or noun.
Similarly, we also obtain negative examples by considering positive examples of other actions that neither share the same Verb nor noun.
More statistics about the dataset are provided in Table \ref{table:stats} and the number of annotated examples per noun (resp.\ verb) is shown in Figure \ref{fig:nouns} (resp. Figure \ref{fig:verbs}).


\section{Evaluation metrics}
\label{metric}

For each action, we collect positives, negatives and hard negatives examples through the annotation system detailed in  Section~\ref{dataset} and illustrated in Figure~\ref{fig:annotation_tool}.
Given these annotations, we can compute the standard mean average precision (mAP) over the different action classes.

One issue when applying the standard mAP metric on this data is that it does not take into consideration that some video contain several positive 10-second video clips of a given action while some others only contain a single positive example.
This creates a bias as video clips coming from the same video are often visually similar and they would tend to weight more in a standard mAP metric.
To address this issue, we instead consider extensions of the mAP that equally weight examples coming from different videos.
We consider the mean weighted average precision scikit-learn implementation\footnote{\url{https://scikit-learn.org/stable/modules/generated/sklearn.metrics.average_precision_score.html}} (mWAP) and weight each video clip with the inverse of the number of annotated video clips coming from the same video.
Alternatively, we also consider computing the standard mAP metric but by subsampling only one annotated video clip per unique video and average the mean subsampled average precision over 100 runs (mSAP n=100).

An evaluation python script for computing these metrics is provided at \projecturl.

\section{Benchmarks}
\label{benchmark}

We provide several benchmarks on \datasetname{} using the S3D text-video model pretrained on HowTo100M\footnote{\url{https://github.com/antoine77340/S3D_HowTo100M}} from \cite{miech2020end}.
Given the pretrained model, a video clip $X$ and a text input $Y$, we can compute the similarity score $s(Y \mid X)$, which measures the relevance of input text $Y$ to video $X$.
Given this model, a video $X$ and an action (Verb, Noun), we can compute $s(Verb \mid X)$, $s(Noun \mid X)$ and $s((Verb, Noun) \mid X)$.

Table \ref{table:benchmark} provides results on the \datasetname{} benchmark using the following baselines: 
$s(Verb \mid X) \times s(Noun \mid X)$ which separately computes the score of the Verb and Noun and combine them in a multiplicative manner (similarly to a logical AND), $s(Verb \mid X) + s(Noun \mid X)$ which also separately computes the score of the Verb and Noun and combine them in a additive manner (similarly to a logical OR) and $s((Verb, Noun) \mid X)$ which jointly models the pair (Verb, Noun).
Surprisingly, $s(Verb \mid X) + s(Noun \mid X)$ performs better than $s(Verb \mid X) \times  s(Noun \mid X)$.

We note that the joint model $s((Verb, Noun) \mid X)$ outperforms the other baselines which suggests that separately detecting either the action verb or the object noun on this benchmark is not a sufficient approach and that more advanced compositionality ability is greatly beneficial.

Including hard negatives for evaluation significantly affects performances which suggests that collecting such negatives is important for the evaluation of compositionality.

\begin{table}[t]
	\centering  
	\begin{tabular}{lccc}
	\toprule
	Method & HNeg & mWAP          & mSAP         \\ \midrule
    $s(Verb \mid X) \times s(Noun \mid X)$ & No & 16.0            & 19.1          \\
	$s(Verb \mid X) + s(Noun \mid X)$      & No & 35.4            & 39.4          \\
	$s((Verb, Noun) \mid X)$               & No & \textbf{40.7}   & \textbf{44.6}          \\ \midrule
    $s(Verb \mid X) \times s(Noun \mid X)$ & Yes & 14.0            & 16.9          \\
	$s(Verb \mid X) + s(Noun \mid X)$      & Yes & 27.0            & 31.0          \\
	$s((Verb, Noun) \mid X)$               & Yes & \textbf{30.5}   & \textbf{34.8}          \\
\end{tabular}
\vspace{1mm}
	\caption{Several baselines using a HowTo100M pretrained model from \cite{miech2020end} on \datasetname. The column HNeg indicates whether or not hard negatives are included at evaluation.}
	\label{table:benchmark}
\end{table}

\section{Conclusion}
In this paper, we have introduced a novel video dataset, \datasetname, annotated with actions involving rare interactions of humans with objects. 
It aims at evaluating the compositionality abilities of action recognition models by combining unlikely pairs of action verbs and objects nouns.
We provided several baselines using a state-of-the-art video and text model and demonstrated that the compositionality ability of the trained model is needed on \datasetname{} to perform well.
We hope the dataset will enable advances in the study of compositionality for action recognition in videos.
\datasetname{} is publicly available for download at \projecturl.

\section{Acknowledgements}
We are grateful to all our additional annotators: Aditya Zisserman and Pauline Métivier.
The project was
partially supported by Antoine Miech Google Ph.D. fellowship.

{\small
\bibliographystyle{ieee_fullname}
\bibliography{master-biblio}
}

\clearpage

\begin{table}
  \centering
  \begin{tabular}{lccc}
	\toprule
Action & Positive & Hard negative & Negative \\ \midrule
blend corn & 6 & 68 & 1762 \\
blend phone & 17 & 197 & 1766 \\
blend pumpkin & 4 & 178 & 1772 \\
blend shoes & 2 & 164 & 1766 \\
cut book & 19 & 303 & 1750 \\
cut car & 30 & 260 & 1736 \\
cut chair & 6 & 396 & 1763 \\
cut coconut & 41 & 335 & 1726 \\
cut keyboard & 22 & 286 & 1745 \\
cut laptop & 29 & 326 & 1738 \\
cut phone & 25 & 495 & 1763 \\
cut pumpkin & 21 & 338 & 1747 \\
cut rock & 10 & 435 & 1759 \\
cut shoes & 22 & 364 & 1744 \\
cut towel & 7 & 257 & 1773 \\
deseed pepper & 9 & 40 & 1758 \\
drill book & 45 & 208 & 1720 \\
drill bottle & 19 & 246 & 1747 \\
drill eggs & 14 & 165 & 1752 \\
drill laptop & 13 & 242 & 1752 \\
drill phone & 32 & 361 & 1733 \\
drill pumpkin & 8 & 254 & 1757 \\
drill rock & 17 & 200 & 1752 \\
drill tomato & 10 & 209 & 1755 \\
drill watermelon & 2 & 275 & 1763 \\
drink chocolate & 2 & 86 & 1782 \\
drink egg & 7 & 97 & 1766 \\
fry phone & 15 & 265 & 1753 \\
hammer banana & 18 & 248 & 1749 \\
hammer bottle & 20 & 294 & 1762 \\
hammer car & 29 & 257 & 1739 \\
hammer coconut & 23 & 336 & 1745 \\
hammer egg & 12 & 266 & 1754 \\
hammer flower & 6 & 263 & 1760 \\
hammer fridge & 2 & 306 & 1764 \\
hammer keyboard & 14 & 273 & 1755 \\
hammer laptop & 12 & 368 & 1753 \\
hammer microwave & 5 & 400 & 1773 \\
hammer phone & 33 & 395 & 1735 \\
hammer pumpkin & 2 & 384 & 1765 \\
hammer shoes & 22 & 403 & 1746 \\
hammer tomato & 13 & 280 & 1753 \\
hammer watermelon & 18 & 369 & 1749 \\
measure chair & 5 & 273 & 1773 \\
measure egg & 13 & 163 & 1752 \\
measure fridge & 21 & 281 & 1781 \\
  \end{tabular}
  \caption{Action classes with their number of collected positive, negative and hard negative samples. Part 1.}
  \label{tab:tax1}
\end{table}
\begin{table}[t]
  \centering
  \begin{tabular}{lccc}
	\toprule
Action & Positive & Hard negative & Negative \\ \midrule
measure hair & 32 & 178 & 1739 \\
measure laptop & 5 & 265 & 1761 \\
measure microwave & 1 & 373 & 1772 \\
measure oven & 12 & 169 & 1759 \\
measure phone & 3 & 330 & 1762 \\
measure pumpkin & 34 & 203 & 1732 \\
measure shoes & 8 & 275 & 1758 \\
measure watermelon & 15 & 249 & 1751 \\
microwave book & 12 & 256 & 1766 \\
microwave bottle & 35 & 217 & 1741 \\
microwave laptop & 6 & 302 & 1760 \\
microwave phone & 25 & 314 & 1740 \\
microwave shoes & 15 & 300 & 1763 \\
microwave watermelon & 15 & 275 & 1751 \\
move towel & 1 & 52 & 1768 \\
open blender & 1 & 118 & 1768 \\
open microwave & 5 & 197 & 1770 \\
peel coconut & 72 & 143 & 1699 \\
peel corn & 6 & 134 & 1763 \\
peel pumpkin & 6 & 204 & 1760 \\
peel watermelon & 3 & 212 & 1762 \\
roll banana & 5 & 136 & 1779 \\
roll carrot & 4 & 113 & 1809 \\
roll potato & 3 & 82 & 1876 \\
roll shirt & 13 & 99 & 1752 \\
shake chair & 1 & 136 & 1767 \\
shake clothe & 3 & 145 & 1762 \\
shake flower & 4 & 47 & 1767 \\
shake hair & 16 & 137 & 1751 \\
shake table & 3 & 145 & 1763 \\
spray banana & 1 & 259 & 1764 \\
spray book & 4 & 336 & 1763 \\
spray chair & 6 & 347 & 1764 \\
spray cup & 9 & 295 & 1798 \\
spray door & 30 & 254 & 1742 \\
spray eggs & 4 & 276 & 1761 \\
spray fridge & 33 & 240 & 1735 \\
spray keyboard & 12 & 273 & 1761 \\
spray laptop & 22 & 308 & 1746 \\
spray microwave & 15 & 459 & 1837 \\
spray phone & 21 & 417 & 1753 \\
spray pumpkin & 23 & 346 & 1749 \\
spray shoes & 38 & 337 & 1741 \\
spray table & 7 & 261 & 1762 \\
throw flower & 7 & 52 & 1764 \\
throw orange & 2 & 68 & 1764 \\
  \end{tabular}
  \caption{Action classes with their number of collected positive, negative and hard negative samples. Part 2.}
  \label{tab:tax2}
\end{table}

\begin{table}[t]
  \centering
  \begin{tabular}{lccc}
	\toprule
Action & Positive & Hard negative & Negative \\ \midrule
throw shirt & 2 & 71 & 1766 \\
unplug fridge & 1 & 87 & 1772 \\
unplug phone & 1 & 278 & 1769 \\
wash apple & 10 & 360 & 1778 \\
wash bicycle & 11 & 307 & 1754 \\
wash blender & 10 & 358 & 1757 \\
wash chair & 55 & 334 & 1715 \\
wash cucumber & 5 & 388 & 1774 \\
wash door & 10 & 385 & 1791 \\
wash fridge & 6 & 394 & 1771 \\
wash keyboard & 14 & 365 & 1755 \\
wash laptop & 20 & 403 & 1745 \\
wash microwave & 22 & 450 & 1756 \\
wash oven & 11 & 356 & 1776 \\
wash pepper & 7 & 363 & 1789 \\
wash potato & 25 & 328 & 1754 \\
wash rock & 26 & 462 & 1739 \\
wash tomato & 20 & 355 & 1755 \\
wash towel & 4 & 415 & 1766 \\
wash watermelon & 25 & 353 & 1741 \\
wash window & 26 & 327 & 1760 \\
weigh banana & 11 & 179 & 1769 \\
weigh book & 21 & 206 & 1747 \\
weigh bottle & 10 & 205 & 1757 \\
weigh egg & 7 & 168 & 1761 \\
weigh phone & 13 & 300 & 1767 \\
weigh pumpkin & 6 & 307 & 1777 \\
weigh shoes & 26 & 241 & 1739 \\
weigh tomato & 26 & 175 & 1781 \\
weigh watermelon & 14 & 229 & 1752 \\
  \end{tabular}
  \caption{Action classes with their number of collected positive, negative and hard negative samples. Part 3.}
  \label{tab:tax3}
\end{table}

\end{document}